# FULL IMAGE RECOVER FOR BLOCK-BASED COMPRESSIVE SENSING

Xuemei Xie*, Chenye Wang, Jiang Du, Guangming Shi

School of Artificial Intelligence, Xidian University

xmxie@mail.xidian.edu.cn

## ABSTRACT

Recent years, compressive sensing (CS) has improved greatly for the application of deep learning technology. For convenience, the input image is usually measured and reconstructed block by block. This usually causes block effect in reconstructed images. In this paper, we present a novel CNN-based network to solve this problem. In measurement part, the input image is adaptively measured block by block to acquire a group of measurements. While in reconstruction part, all the measurements from one image are used to reconstruct the full image at the same time. Different from previous methods recovering block by block, the structure information destroyed in measurement part is recovered in our framework. Block effect is removed accordingly. We train the proposed framework by mean square error (MSE) loss function. Experiments show that there is no block effect at all in the proposed method. And our results outperform 1.8 dB compared with existing methods.

## 1. INTRODUCTION

Compressive sensing (CS) theory shows that signal can be reconstructed under an extremely low sample rate because of its sparse structure. In conventional CS problem, signal is measured with Gaussian matrix in blocks and then recovered by optimization algorithms [1] [2] [3] [4] [5] [6].

Recently, CNN-based methods [7] [8] [9] [10] are proposed for CS problem. The network can adaptively learn a transform from measurements to reconstruction images by minimizing error between the original and the reconstructed images in large dataset. When it comes to testing, the speed and accuracy can be greatly improved. Most existing methods use block-wise approach, in which the input images are measured and recovered block by block [7] [8] [9] [11] [12]. It is convenient to implement. In addition, by segmenting images into blocks, images with any size can be measured. However, in these methods the blocks are reshaped into columns [13], which damage the structure information and result in serious block effects especially in low measurement rate. Thus, denoising algorithms such as BM3D [14] are usually used to remove it. The method proposed in [15] firstly uses fully convolutional network to measure the full image

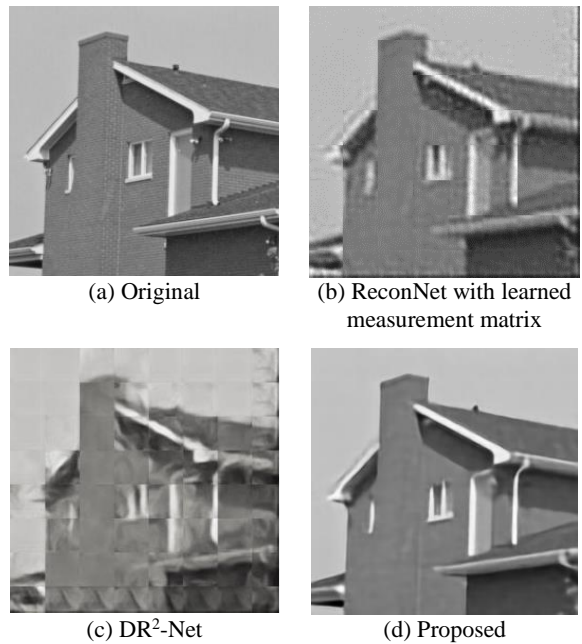

(a) Original  (b) ReconNet with learned measurement matrix

(c) DR$^2$-Net  (d) Proposed

**Fig. 1.** The figure shows the results recovered by ReconNet with learned measurement matrix, DR$^2$-Net and proposed method for measurement rate 4%.

directly. Because of the overlapped structure, the block effect is suppressed effectively in the measurement part.

In this paper, we propose a novel CS framework which removes the block effect efficiently as is shown in Fig.1. In the measurement part, the input image is adaptively measured block by block, which can be easily implemented on optical system. Different from pervious methods, in reconstruction part, we employ resnet [16] to recover the full image from its block-wise measurements. During the reconstruction stage, block effect is also removed. By minimizing the error between the original and the reconstructed images, the network can be trained end-to-end. Experimental results show the excellent performance of proposed framework on both gray images and color images recovery.

The organization of this paper is as follows. Section 2 introduces some related work about CNN-based CS. Section 3 describes the detailed architecture of the proposed method. Section 4 presents the experimental results of the proposed method compared to typical existing methods. Section 5 draws the conclusion of our work.

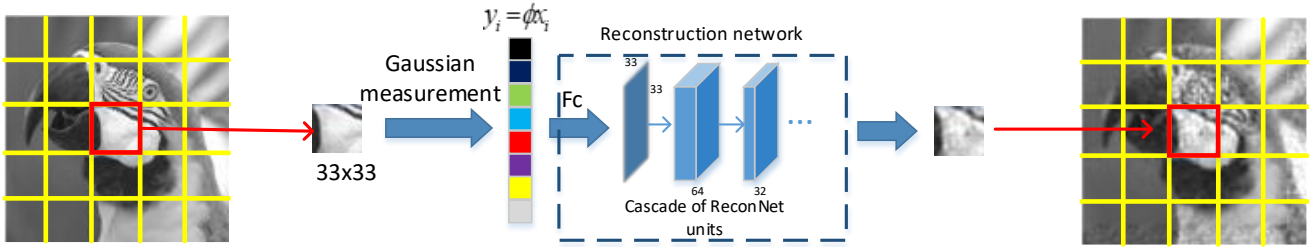

**Fig. 2.** The framework of ReconNet. The scene is measured by Gaussian matrix block by block. And then the measurements are send to Reconnet units to acquire reconstructed blocks

## 2. RELATED WORK

### 2.1. CNN-based CS

In recent years, deep neural network is widely applied to CS problem. Mousavi, Patel, and Baraniuk [10] firstly apply deep neural network to CS problem. ReconNet [7] and DeepInverse [17] train convolutional neural networks to recover signals from measurements. Based on ReconNet, $DR^2$-Net [8] obtains an improved result by learning the residual information between low resolution images and ground truth images. Adp-Rec [11], DeepCodec [12] and new ReconNet [9] jointly train the measurement and the reconstruction parts. They acquire excellent performance. Xie *et. al.* [15] firstly proposed fully convolutional measurement network where the input is measured by overlapped convolution operation. This method removes the block effect effectively.

### 2.2. Block by block recovery

In most CNN-based methods [7] [9] [11] [12], the input is usually measured and recovered block by block. That is because measuring a full image directly needs a lot of computer memory. Moreover, block-based CS can be easily implemented by optical devices [7] [9] [18] [19] [20]. The architecture of ReconNet is shown in Fig. 2. The image block is firstly measured by Gaussian matrix. Then a fully connected layer is employed to get a low resolution image. After that, the ReconNet units is used to produce reconstruction image block. The loss function is given by

$$L(\{W\}) = \frac{1}{T}\sum_i^T \|f(y_i, \{W\}) - x_i\|_2^2 \qquad (1)$$

where $f(y_i, \{W\})$ is the $i$−th reconstructed image block of ReconNet. And $x_i$ is the $i$–th original image block as well as the $i$−th label. $\{W\}$ means the training parameters in ReconNet. T is the total number of image blocks in the training dataset. In this method, since all the blocks are independent in both measurement and reconstruction parts, there is serious block effect in reconstructed image, especially at low measurement rates. The latest CS network [15] measures the scene by overlapped convolutional operation. It can remove block effect successfully, while the implementation is still in exploration.

## 3. PROPOSED METHOD

This paper proposes a block-based measurement and full image reconstruction network. It can be implemented by existing method. Meanwhile it can achieve pleasant visual experience with no block effect absolutely.

### 3.1. Network architecture

The architecture of the proposed framework is shown in Fig.3. The network is composed by measurement part and reconstruction part. The measurements are obtained by non-overlapped convolution operation. This is same as traditional block-based measurement. All the kernels make up an adaptive measurement matrix. They measure the input images block by block with sliding windows. It is easy to be implemented by the normally used optical devices.

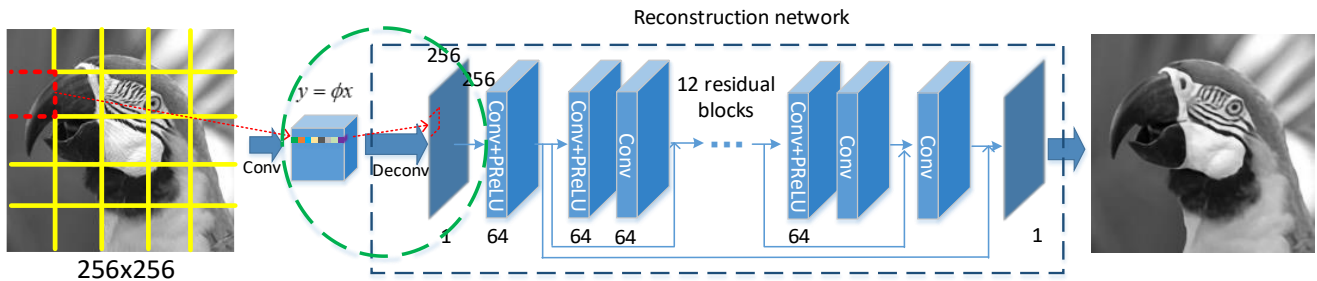

**Fig. 3.** Proposed CS architecture. The network composed by measurement part and reconstruction part. The scene is measured block by block, while the reconstruction network recovers the full image from its measurements.

Different from block by block reconstruction, all the measurements are used to recover the image at the same time. As is shown in green dotted line in Fig.3. While the previous work is recovered block by block as is shown in Fig.2. In reconstruction phase, a deconvolutional layer is firstly used to produce a low resolution reconstruction image. Inspired by [21] [22] [16], we employ residual blocks in reconstruction network. Each residual block contains two convolution layers and a PReLU activation function. The detailed structure is shown in Fig.3.

Accordingly, the loss function of proposed framework is given by

$$L(\{W\}) = \frac{1}{T}\sum_{i=1}^{T}\left\|f\left(I_i^{hr},\{W\}\right) - I_i^{hr}\right\|_2^2 \quad (2)$$

where $\{W\}$ represents the parameters of proposed network. T is the total number of full images in the training batch. Different between (1) and (2) is that in (2) the input is full image $I_i^{hr}$ but block measurement $y_i$ in (1). By minimizing the mean square error (MSE) loss function, the network can be trained end to end.

### 3.2. Full image recover and deblock

Block-based measurement usually destroy the structure information. After reconstructing block by block, all the image blocks are spliced together stiffly. In fact, there is no any connection among these blocks. And block effect appears

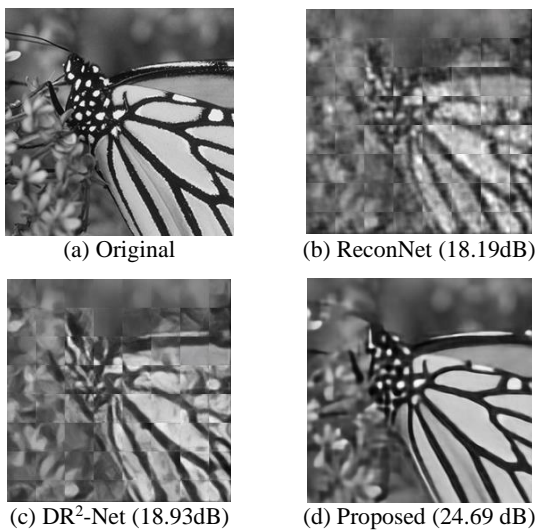

(a) Original      (b) ReconNet (18.19dB)

(c) DR$^2$-Net (18.93dB)      (d) Proposed (24.69 dB)

**Fig. 4.** The reconstruction result of proposed method compared with RceconNet and DR$^2$ for measurement rate 4%

accordingly. As is shown in fig4 (b) and (c). In the proposed framework, however, the reconstruction network deals with all the measurements as a whole. Because of convolution operation whose kernel size and stride is small, the connection of block measurements is rebuild. In addition, the loss function of the proposed method guide reconstruction network to minimize the error of full images between the input and the output instead of image blocks. The obvious additional information in reconstruction images such as block effect is not tolerated. So, the recovery network remove block effect at its reconstruction end.

The compared results of ReconNet, DR2-Net and the proposed method with measurement rate 4% are shown in Fig.4. It is obviously that our results have no block effect absolutely even if the measurement rate is very low. And the structure of the image is completely preserved. This makes the visual effect excellent. In addition, the PSNR is much higher than the other two methods.

## 4. EXPERIMENTS

In this section, we conduct the experiments on MSE loss with fully convolutional neural network for compressive sensing problem.

We use the tensorflow [23] framework for network training and testing. Our computer is equipped with Intel Core i7-6700 CPU with frequency of 3.4GHz, 4 NVidia GeForce GTX Titan XP GPUs, 128 GB RAM, and the framework runs on the Ubuntu 16.04 operating system. The training dataset consists of 800 images from DIV2K dataset. Train images are obtained by cropping random $256 \times 256$ high resolution images from train dataset. Before entered into the network, these images are converted to grayscale and are scaled the pixel values to $[-1,1]$. All the models at different measurement rate are trained with learning rate of $10^{-4}$ and 1000 epochs.

We compare the proposed framework with other three classical method ReconNet, DR$^2$-Net and Adp-Rec respectively. Some examples of test images are shown in Fig.5. The proposed network is trained with measurement rate of 1%, 4%, 10% and 25%. The detailed compared results with four measurement rates is shown in Table.1. It obviously that the proposed method acquired best performance on both PSNR and visually effects. The block effects are removed clearly even if at low measurement rate.

From Fig.5 (a) with measurement rate of 1%, it is clear that ReconNet, DR$^2$-Net and Adp-Rec have serious block effects. It is even hard to distinguish the content in the first two reconstructed images. In contrast, the results of the proposed method have no any block effect even in such a low measurement rate. When measurement is 4% in Fig.5 (b), the block effect in first two results is obviously in high-frequency areas and the edges of reconstruction images. The result of Adp-Rec is better than the first two methods, but the block effect in eyes of 'Lena' influence the visual effect seriously. In Fig.5 (c), the results of Adp-Rec and the proposed method surpass the other two methods clearly. Though the block effect in Adp-Rec is weaken, there is still much noises compared with proposed method because of block recovery. When the measurement is 25%, there is no obvious block effect in all methods. While the proposed method have best visual experience.

The detailed comparing results is show in Table. 1. It is obviously that the results of our method acquire the best performance. The PSNR is higher than the state-of-the-art method by 1.8 dB on average. The superiority is especially clear at the measurement rate 25%

## 5. CONCLUSION

This paper proposes a novel framework of CS where the measurements are measured block by block. Different from the previous framework, the full image is reconstructed at one time in our work. In addition, the proposed method is easier for implementation, and outperforms the state-of-the-art methods. In the future, we will apply the proposed method on hardware and set up a real-time compressive sensing camera with FPGA.

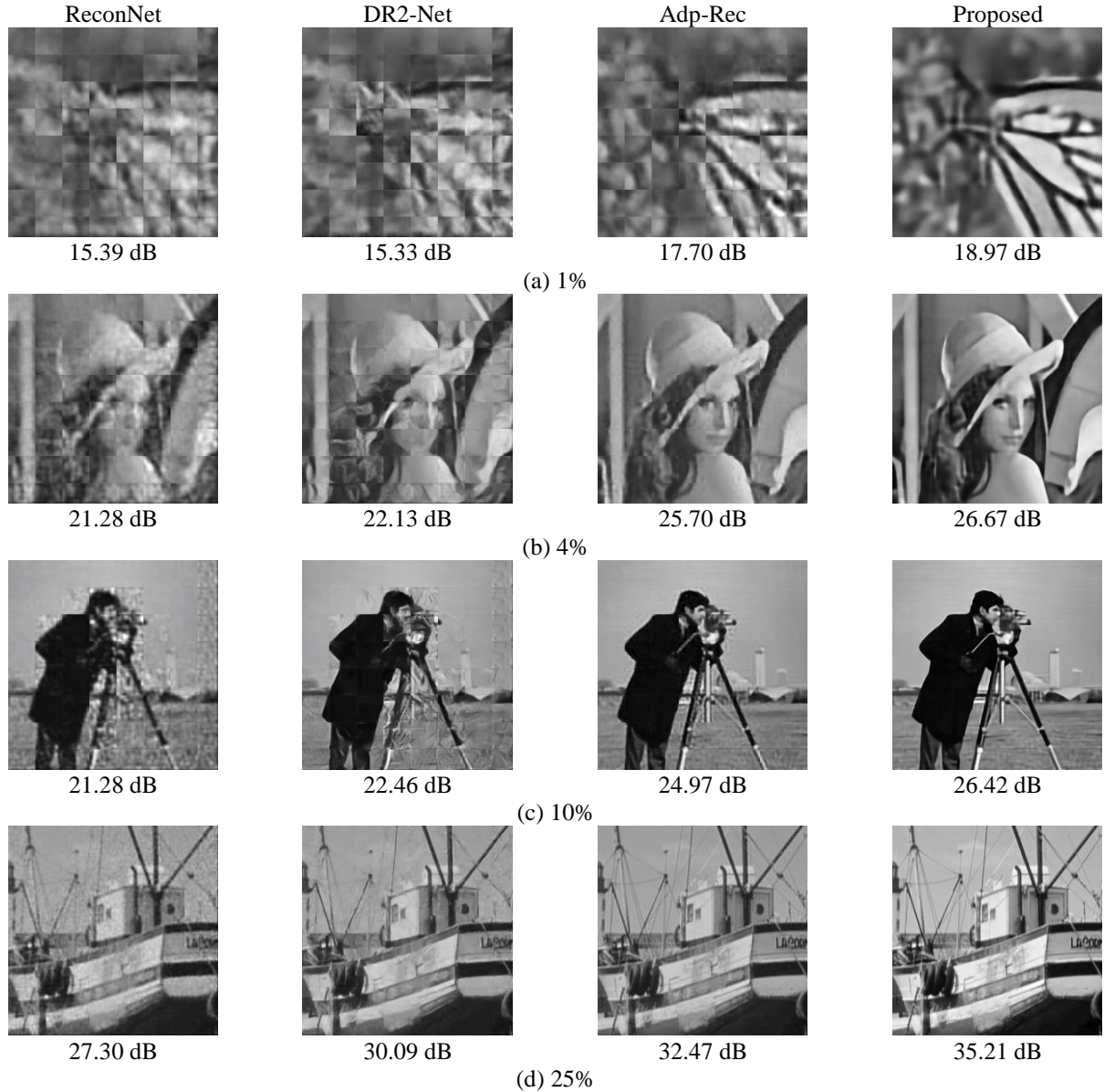

**Fig. 5.** The reconstruction results for measurement rate 1%, 4%, 10% and 25%.

| | Image Name | ReconNet | DR$^2$-Net | Adp-Rec | Proposed |
|---|---|---|---|---|---|
| **1%** | Monarch | 15.39 dB | 15.33 dB | 17.70 dB | 18.97 dB |
| | Parrots | 17.63 dB | 18.01 dB | 21.67 dB | 22.68 dB |
| | Barbara | 18.61 dB | 18.65 dB | 21.36 dB | 22.13 dB |
| | Boats | 18.49 dB | 18.67 dB | 21.09 dB | 22.28 dB |
| | Cameraman | 17.11 dB | 17.08 dB | 19.74 dB | 20.92 dB |
| | House | 19.31 dB | 19.61 dB | 22.93 dB | 24.71 dB |
| | Lena | 17.87 dB | 17.97 dB | 21.49 dB | 23.19 dB |
| | **Mean** | **17.77 dB** | **17.90 dB** | **20.85 dB** | **22.12 dB** |
| **4%** | Monarch | 18.19 dB | 18.93 dB | 22.91 dB | 24.69 dB |
| | Parrots | 20.27 dB | 21.16 dB | 24.35 dB | 25.50 dB |
| | Barbara | 20.38 dB | 20.70 dB | 23.42 dB | 23.76 dB |
| | Boats | 21.36 dB | 22.11 dB | 25.29 dB | 27.13 dB |
| | Cameraman | 19.26 dB | 19.84 dB | 22.54 dB | 23.78 dB |
| | House | 22.58 dB | 23.92 dB | 27.75 dB | 30.29 dB |
| | Lena | 21.28 dB | 22.13 dB | 25.70 dB | 26.67 dB |
| | **Mean** | **20.47 dB** | **21.26 dB** | **24.57 dB** | **25.97 dB** |
| **10%** | Monarch | 21.10 dB | 23.10 dB | 26.65 dB | 29.24 dB |
| | Parrots | 22.63 dB | 23.94 dB | 27.59 dB | 28.75 dB |
| | Barbara | 21.89 dB | 22.69 dB | 24.28 dB | 24.50 dB |
| | Boats | 24.15 dB | 25.58 dB | 28.80 dB | 30.56 dB |
| | Cameraman | 21.28 dB | 22.46 dB | 24.97 dB | 26.42 dB |
| | House | 26.69 dB | 27.53 dB | 31.43 dB | 33.42 dB |
| | Lena | 23.83 dB | 25.39 dB | 28.50 dB | 29.71 dB |
| | **Mean** | **23.08 dB** | **24.38 dB** | **27.46 dB** | **28.94 dB** |
| **25%** | Monarch | 24.31 dB | 27.95 dB | 29.25 dB | 34.36 dB |
| | Parrots | 25.59 dB | 28.73 dB | 30.51 dB | 34.00 dB |
| | Barbara | 23.25 dB | 25.77 dB | 27.40 dB | 29.16 dB |
| | Boats | 27.30 dB | 30.09 dB | 32.47 dB | 35.21 dB |
| | Cameraman | 23.15 dB | 25.62 dB | 27.11 dB | 30.40 dB |
| | House | 28.46 dB | 31.83 dB | 34.38 dB | 37.45 dB |
| | Lena | 26.54 dB | 29.42 dB | 31.63 dB | 34.42 dB |
| | **Mean** | **25.51 dB** | **28.49 dB** | **30.39 dB** | **33.57 dB** |

Table. 1. Results of the proposed method and classical methods at different measurement rates